\documentclass[10pt,twocolumn,letterpaper]{article}

\usepackage{cvpr}
\usepackage{times}
\usepackage{epsfig}
\usepackage{graphicx}
\usepackage{amssymb}
\usepackage{epstopdf}

\usepackage{cite}
\usepackage{comment}
\usepackage{tabularx}
\usepackage{array}
\usepackage{booktabs}
\usepackage{caption}
\usepackage{bm}
\usepackage{subcaption}
\usepackage{float}
\usepackage{color}
\usepackage[cmex10]{amsmath}
\usepackage[pagebackref=true,breaklinks=true,letterpaper=true,colorlinks,bookmarks=false]{hyperref}

\DeclareMathOperator*{\argmin}{arg\,min}

\newif\ifcomments
\commentstrue

\ifcomments
    \usepackage{ulem}
    \def\mc#1{{\color{blue} [\textbf{MC:} #1]}}
    
    \def\jlk#1{{\color{red} [\textbf{JLK:} #1]}}
    
    \def\piedit#1{{\color{magenta} [\textbf{PI:} #1]}}
    
\else 
    \def\mc#1{}
    
    \def\jlk#1{}
    
    \def\piedit#1{}
    
\fi

\usepackage[pagebackref=true,breaklinks=true,letterpaper=true,colorlinks,bookmarks=false]{hyperref}

\cvprfinalcopy 

\def\cvprPaperID{7} 

\ifcvprfinal\fi
\begin{document}


\title{Low-Resolution Overhead Thermal Tripwire for Occupancy Estimation}

\author{Mertcan Cokbas, Prakash Ishwar, Janusz Konrad\\
Boston University\\
8 Saint Mary's Street, Boston, MA 02215\\
{\tt\small [mcokbas, pi, jkonrad]@bu.edu}}


\maketitle

\begin{abstract}
   Smart buildings use occupancy sensing for various tasks ranging from energy-efficient HVAC and lighting to space-utilization analysis and emergency response. We propose a people counting system which uses a low-resolution  thermal sensor. Unlike previous people-counting systems based on thermal sensors, we use an overhead tripwire configuration at entryways to detect and track transient entries or exits. We develop two distinct people counting algorithms for this configuration. To evaluate our algorithms, we have collected and labeled a low-resolution thermal video dataset using the proposed system. The dataset, the first of its kind, is public and available for download\footnote{\href{http://vip.bu.edu/tidos}{\tt vip.bu.edu/tidos}}. We also propose new evaluation metrics that are more suitable for systems that are subject to drift and jitter.
\vglue -0.3cm
\end{abstract}

\section{Introduction}

Occupancy sensing is a key technology for smart buildings of the future \cite{Thermosense}, \cite{IoT}, \cite{Survey_Energy_Efficiency}. The knowledge of where and how many people are in a building enables, among others, smart HVAC control to save energy, space management to reduce rental costs and enhanced security, (e.g., fire, flooding, active shooter) \cite{Multi-sensing_paper}. Over the years, several people-counting systems have been proposed leveraging various sensing modalities, e.g., surveillance cameras, MAC address trackers, WiFi signal measurement, CO${_2}$ sensors and thermal sensors \cite{comparison_paper},\cite{wifi_tracking}, each with its own deficiencies. For instance, surveillance cameras may not be acceptable in scenarios where privacy is expected, MAC address trackers require people to carry a networked device, WiFi signal measurement is sensitive to EM interference and unreliable for crowds of people, while CO${_2}$ sensors have delayed reaction times due to slow mixing of gases. On the other hand, thermal sensors do not suffer from any of these issues. 

To date, people-counting methods using low-resolution thermal sensors have focused on assessing the state of a room's interior \cite{Thermosense}, \cite{Tyndall_paper}, \cite{Elevator_paper}. Such methods can be effective for small rooms, but in case of a large room the field of view (FOV) of a low-resolution thermal sensor might not be sufficient to capture all people in the room. In this scenario, multiple sensors are needed but this increases the cost, complexity of installation and requires complex processing to avoid overcounts due to FOV overlap.
%
%
\begin{figure}[t!]
\centering
    \centering
    \includegraphics[scale=0.3]{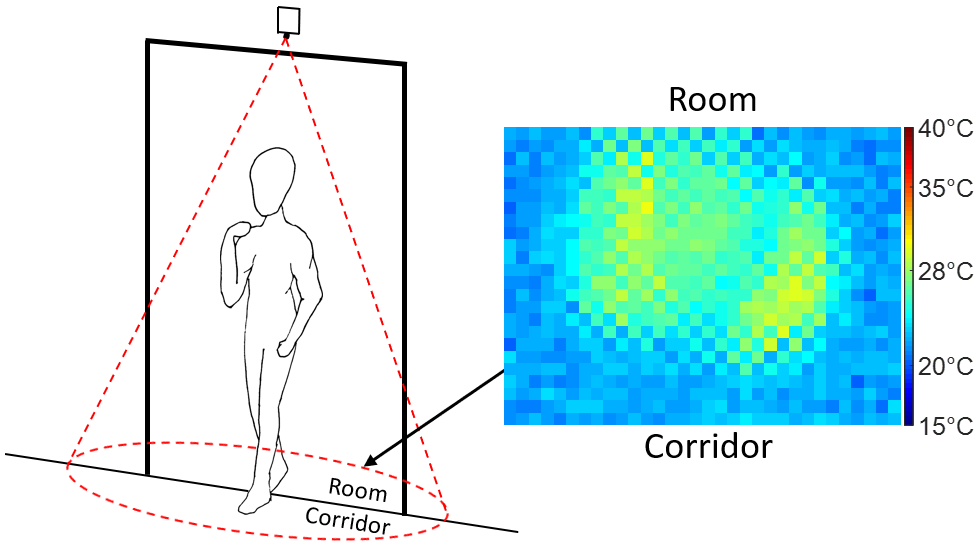}
    \label{fig:front_view}
    \caption{Configuration of our virtual-tripwire door setup: low-resolution thermal sensor mounted above a door and facing down (left); and 32$\times$24-pixel thermal frame captured by the sensor when a person is leaving the room (right).}
    \label{fig:configuration_sketch}
    \vglue -0.5cm
\end{figure}

In contrast, we propose to count people using a single low-resolution thermal sensor mounted above {\it every} entry/exit point of a room (Fig.~\ref{fig:configuration_sketch}) and develop a computational methodology to accomplish this. Regardless of room size, such thermal tripwires can independently detect people entering/exiting a room and jointly estimate the occupancy (state) of the room. In contrast to past methods, our approach is not frame-based but event-based, that is a people count is updated only upon the completion of a door event.

The approach we propose consists of three steps: background subtraction, event detection and event classification. In the first step, we detect ``warm'' pixels {\it via} a probabilistic background-temperature model based on Running Gaussian Average \cite{running_gaussian_average}. Since this model does not leverage spatial coherence of temperature, we combine it with a Markov Random Field (MRF) model \cite{MRF} to produce high-temperature blobs. In the second step, based on background/foreground separation, we detect door events. In a baseline version, we assume that one person passes through the door at a time and we treat all foreground pixels as associated with this person.
In order to handle wider doors and multiple people, we develop an enhanced algorithm that identifies high-temperature blobs and tracks them. In the third step, we classify each event as an entry or exit based on the direction of blob movement. To validate the performance of our algorithms, we have collected and manually labeled a dataset of thermal sequences covering various scenarios, including challenging edge cases. This dataset, the first of its kind, is public and available for download. We evaluate our algorithms on this dataset and show that while both proposed algorithms work equally well in normal scenarios the enhanced algorithm outperforms the baseline algorithm on edge cases.

To summarize, our paper makes several contributions. First, we believe this is the first work to develop and systematically study the overhead virtual tripwire configuration of a low-resolution thermal sensor for people counting. Second, we develop and validate two distinct people-counting algorithms based on low-resolution thermal data that are capable of handling challenging edge cases. Third, this work makes available to the research community a dataset of thermal sequences with manual annotations of entries into and exits from a room. 
Finally, we propose new metrics that provide a more meaningful evaluation of performance for systems that suffer from jitter and cumulative errors due to drift.

\section{Related Work}

Erickson {\it et al.} \cite{POEM_paper} proposed a people-counting system, called POEM, for energy management in buildings using a combination of  video cameras in hallways and PIR sensors in rooms. Data coming from the camera and PIR networks are fused to estimate the people count.

ThermoSense \cite{Thermosense} is the first system, to the best of our knowledge, that uses thermal sensors for people counting. However, unlike in our thermal-tripwire approach, each of its sensor's FOV must cover the whole room to ``see'' all room occupants.
ThermoSense estimates occupancy for each frame in three main steps: background subtraction, feature extraction, and people-count classification. 
The background subtraction is similar to that of ours; each pixel's temperature is modeled by mean and standard deviation.
A set of features are extracted from foreground/background segmentation, namely: the number of foreground pixels, the number of connected components, and the size of the largest connected component. These features are used to estimate occupancy by means of linear regression, kNN classification or neural network. In addition to thermal sensors, ThermoSense uses PIR sensors to detect whether a room is empty or not. This helps smooth out raw estimates and update the background model. Tyndall {\it et al.} \cite{Tyndall_paper} proposed some improvements to ThermoSense. They showed that entropy-measure classifiers such as K* and C4.5 work better for their use cases. However, the sensors that are used in these two studies are different. Thermosense uses an $8 \times 8$ Grid-Eye sensor, whereas Tyndall {\it et al.} \cite{Tyndall_paper} use a $16 \times 4$ Melexis sensor. 

Amin {\it et al.} \cite{Elevator_paper} proposed a people-counting system that uses a video camera and a thermal sensor both pointed inside a room, unlike in our approach. The use of two imaging modalities is meant to improve system robustness, e.g., in case of illumination variations. The counts are estimated separately using camera images and thermal frames and then linearly combined into a final people-count estimate.

\section{Methodology}

\begin{figure}[h!]
\centering
            \includegraphics[scale=0.3]{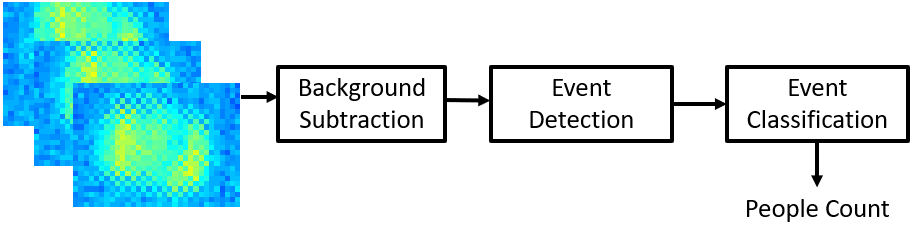}
            \caption{Block diagram of the proposed approach.
            }
            \centering
          \label{fig:main block diagram}
\vglue -0.2cm
\end{figure}

In our approach, we analyze consecutive thermal frames captured by a sensor mounted above a door (Fig.~\ref{fig:configuration_sketch}) in three steps: (1) background subtraction to first detect the presence of one or more people in the FOV of the sensor; (2) event detection to identify the beginning and end of entry or exit events spanning multiple frames; and (3) event classification as an entry or exit (Fig.~\ref{fig:main block diagram}). These three steps are discussed in detail below.

\subsection{Background Subtraction}

\begin{figure}[h!]
\centering
    \begin{subfigure}[h]{0.47\textwidth}  
    \centering
            \includegraphics[width=0.7\textwidth]{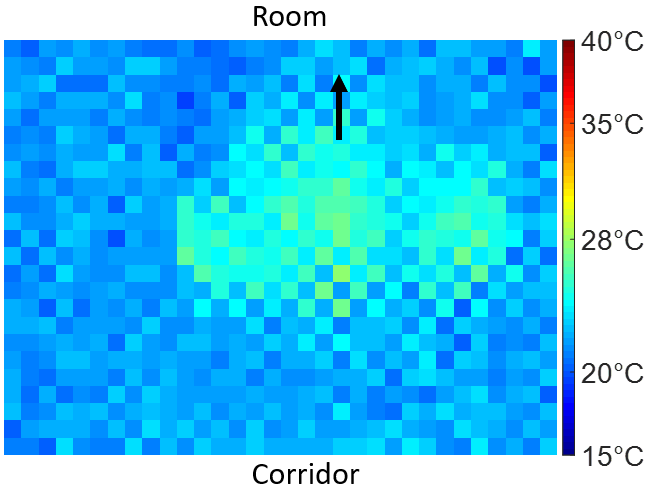}
            \caption{32$\times$24-pixel frame from Melexis MLX90640 sensor with person passing through a door. Rows of the frame are aligned with the door frame while columns are orthogonal to the door opening.}
            \label{fig:sample_thermal}
    \end{subfigure}\quad
    \begin{subfigure}[h]{0.47\textwidth}
            \centering
            \includegraphics[width=0.7\textwidth]{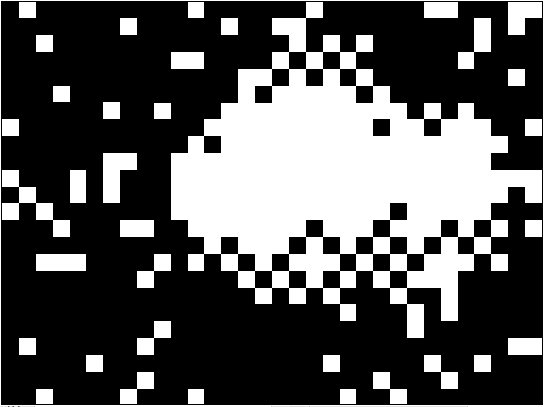}
            \caption{Result of background subtraction using RGA BS algorithm.}
            \label{fig:no_MRF}
    \end{subfigure}\quad
    \begin{subfigure}[h]{0.47\textwidth}
            \centering
            \includegraphics[width=0.7\textwidth]{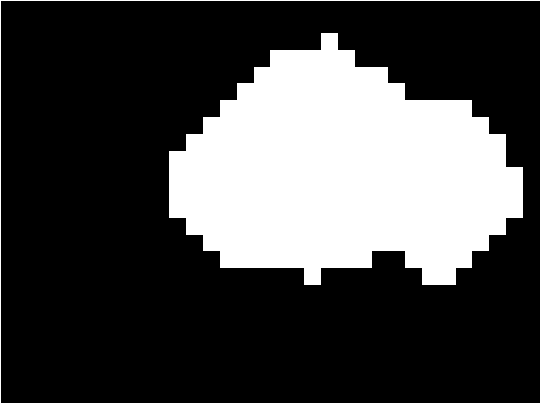}
            \caption{Result of background subtraction using RGA+MRF BS algorithm.}
            \label{fig:with_MRF}
    \end{subfigure}
    \caption{Thermal frame and results of background subtraction for a single person passing through a door.}\label{fig:MRF}
\end{figure}

In this step, our goal is to separate the pixels that correspond to a human body from those that belong to the background (floor, walls, other surroundings). Since the system is designed for indoor people counting, it is reasonable to assume that a person is warmer than the background. Despite the difference between body temperature and room temperature, a single global threshold cannot reliably distinguish between them due to natural variations in people and indoor environments. In our approach, instead of thresholding temperature values, we model the background temperature of each pixel by a Gaussian pdf and apply a threshold to the temperature \textit{probabilities}.  Let \textit{T}$_n$[\textbf{$\bm{x}$}] denote the temperature value of a pixel at location \textbf{$\bm{x}$} in frame \textit{n}.
We use the Running Gaussian Average (RGA) model \cite{BS_review}, \cite{Adaptive_BGS} to update the mean \textit{$\mu$}$_n$[\textbf{$\bm{x}$}] at every background location $\bm{x}$ as follows:
\begin{multline}
  \mu_n[\bm{x}] = \bm{1}(T_n[\bm{x}]\epsilon B) \Big [\alpha T_n[\bm{x}] + (1 - \alpha) \mu_{n-1}[\bm{x}]) \Big] \\ + \bm{1}(T_n[\bm{x}]\epsilon F)\mu_{n}[\bm{x}]
\end{multline} 
where the sets of background and foreground pixels are denoted by \textit{B} and \textit{F}, respectively, $\bm{1}(\cdot)$ is an indicator function, and $0<\alpha<1$ is a weight controlling recursive update of the mean. We model the probability that a pixel at $\bm{x}$ belongs to the background as follows:
\begin{equation}
  {P_{B}(T_n[\bm{x}])} = \mathcal{N}(T_n[\bm{x}]-\mu_n[\bm{x}], \sigma)
\end{equation} 
where $\mathcal{N}(\cdot,\cdot)$ denotes the Gaussian distribution with standard deviation $\sigma$. We use the same fixed $\sigma$ for all pixels and perform background subtraction by means of the following binary hypothesis test applied to $P_B(\cdot)$:
\begin{equation}
  {P_{B}(T_n[\bm{x}])} \underset{{F}}{\overset{B}{\gtrless}} \eta
\end{equation} 
where $\eta$ is a fixed threshold, identical for all pixels. We refer to this overall background subtraction model as Running Gaussian Average based Background Subtraction (RGA BS) and show a sample result in Fig.~\ref{fig:no_MRF}.
%

The background subtraction model discussed so far uses temporal information to separate the foreground from the background. However, all decisions are made independently at neighboring pixels, thus leading to fragmented body-temperature areas. In order to address this, one needs to leverage the spatial contiguity of the human body by applying spatial constraints to foreground estimates. For this purpose, we use an approach proposed by McHugh {\it et al.} \cite{MRF}. They used a Markov Random Field (MRF) model to ensure spatial estimate coherence within a binary hypothesis test as follows:
\begin{equation}\label{eqn:binhyptest}
  \frac{P_{B}(T_n[\bm{x}])}{P_{F}(T_n[\bm{x}])} \underset{{F}}{\overset{B}{\gtrless}}
  \theta exp\Big(\frac{Q_{F}[\bm{x}]-Q_{B}[\bm{x}]}{\gamma}\Big),
\end{equation} 
where $P_{F}(T_n[\bm{x}])$ is the probability that $T_n[\bm{x}]$ belongs to the foreground, $Q_{F}[\bm{x}]$ and $Q_{B}[\bm{x}]$ denote the number of neighboring foreground and background pixels around location $\bm{x}$, respectively, while $\theta$ and $\gamma$ are parameters. Unlike $P_{B}(\cdot)$,  we assume $P_{F}(\cdot)$ is a constant (uniform distribution) because we observed that the foreground (body) temperature footprint characteristics can vary significantly depending on clothing, hairstyle and height of a person. Effectively, the right-hand side of the binary hypothesis test (\ref{eqn:binhyptest}) is a spatially-adaptive threshold. Depending on the labels of neighboring pixels, the threshold will change. If there are more foreground pixels than background pixels in the neighborhood of $\bm{x}$, the threshold will increase, and, therefore, it will be more likely that the pixel is deemed as belonging to the foreground (and vice versa). Due to the variable threshold, the MRF model increases spatial coherence of foreground estimates, which can be seen in Fig.~\ref{fig:with_MRF}. The parameter $\gamma$ can be used to adjust the degree to which the MRF model impacts the threshold.

\subsection{Event Detection}

We propose two different event detection algorithms.
Our baseline  algorithm assumes that no more than one person will pass under a door at a given time. Our multi-person algorithm, however, is designed to handle multiple people simultaneously passing through the door.



\subsubsection{Baseline Algorithm}


We define an event as a sequence of consecutive frames that satisfy the following conditions: (1) the frames immediately preceding and following the event are empty, i.e., have no foreground pixels, (2) each frame in the event has at least one foreground pixel, and (3) there is at least one frame in the event with at least $K$ foreground pixels, were $K$ is a parameter which can be adjusted to account for the  height at which the sensor is mounted above the door (smaller $K$ for greater heights).

\begin{figure}[h!]
\centering
    \begin{subfigure}[h]{0.47\textwidth}  
    \centering
            \includegraphics[width=0.7\textwidth]{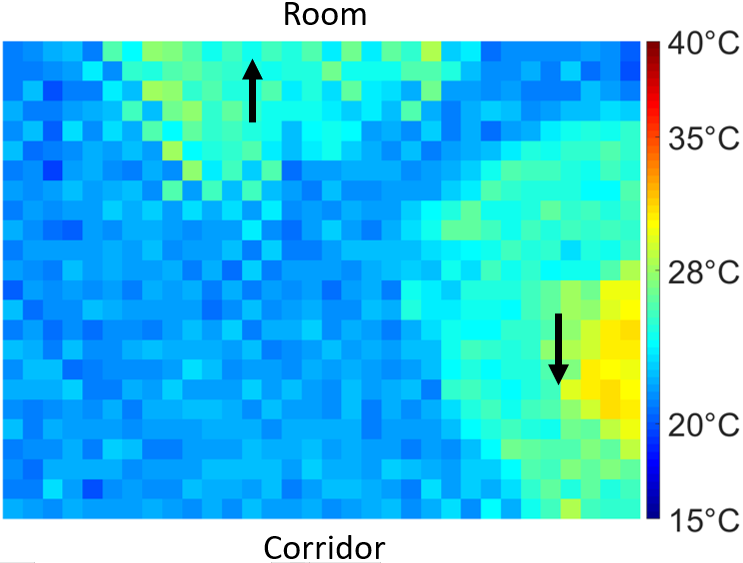}
            \caption{32$\times$24-pixel frame with two people passing through a door. 
            }
            \label{fig:two people orig}
    \end{subfigure}\quad
    \begin{subfigure}[h]{0.47\textwidth}
            \centering
            \includegraphics[width=0.6\textwidth]{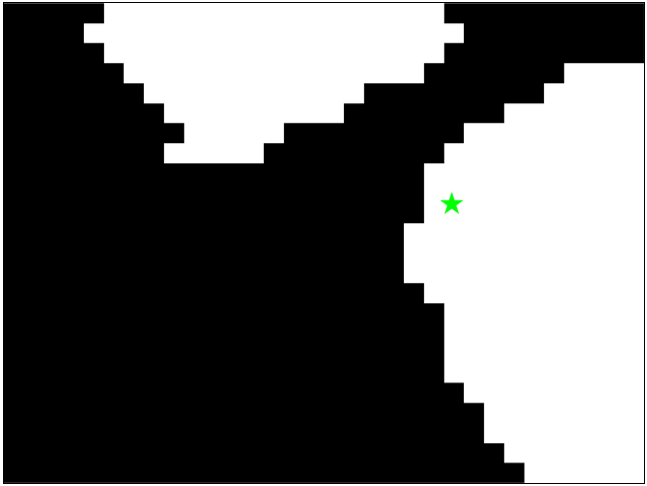}
            \caption{Result of background subtraction using RGA+MRF BS algorithm with centroid (green star) computed using the baseline algorithm.}
            \label{fig:two people single}
    \end{subfigure}\quad
    \begin{subfigure}[h]{0.47\textwidth}
            \centering
            \includegraphics[width=0.6\textwidth]{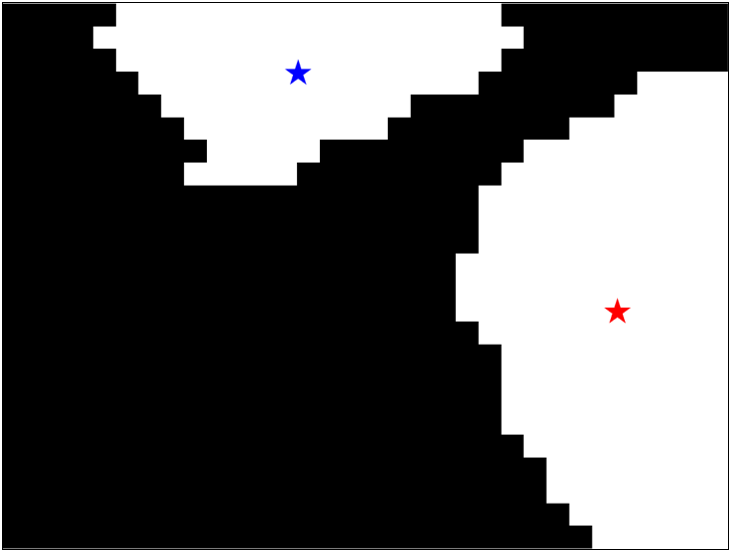}
            \caption{Result of background subtraction using RGA+MRF BS algorithm with two centroids (red and blue stars) computed using the multi-person algorithm.}
            \label{fig:two people multiple}
    \end{subfigure}
    \caption{Thermal frame and results of background subtraction and centroid calculation for 2 people passing through a door.}\label{fig:Comparison of Events}
\end{figure}

\subsubsection{Multi-Person Algorithm}

In the baseline algorithm, we assumed that only one person passes under the sensor at a time. If multiple people pass through the door within the same event, the algorithm is incapable of distinguishing them (it calculates only one centroid), thus resulting in an error (Fig.~\ref{fig:Comparison of Events}).


To address this, we detect blobs among foreground pixels in each frame and track their movement. A {\it blob} is defined as a connected component of foreground pixels of size $L$ or more. We also define a {\it blob track} as a time sequence of blobs, one in each frame, that are linked between consecutive frames {\it via} association described below. We consider each blob track to be an event.  Blob tracks start, grow and end as described below.

\textit{Blob track birth:} If there are more blobs in the current frame than in the previous frame, a new blob track is created. The decision as to which blob will be associated with the new blob track is determined after data association in the growth phase.

\textit{Blob track growth:} If the number of blobs in the current and previous frames is the same, then a one-to-one mapping is established between blobs in those frames thus leading to track growth. The track to which a previous-frame blob belongs is grown by a current-frame blob with which the previous-frame blob is associated. This association is established based on the Euclidean distance between blobs' centroids. First, for each blob in the current frame the closest blob is found in the previous frame. Then, the blob pair with the smallest centroid-to-centroid distance is said to be associated with each other and removed from further consideration. The procedure is repeated for the remaining current-frame blobs. Other blob association methods could be applied as well, e.g., minimization of the sum of distances for all blob pairs. However, sophisticated methods may not work as well in our application context due to low thermal sensor resolution, short duration of events and the similarity of thermal footprints of different people.

\textit{Blob track termination:} If there are fewer blobs in the current frame than in the previous frame, a blob track is terminated. The decision as to which blob is to be terminated is determined after data association in the growth phase.

\subsection{Event Classification}

Both algorithms classify each event at its completion into one of the following classes: (1) a person left the room or (2) a person entered the room. This is accomplished by analyzing the direction of movement of foreground pixels throughout the event. Let $F_n$ be defined as follows:
\begin{itemize}
    \item baseline algorithm: a set of all foreground pixels at time $n$,
    \item multi-person algorithm: a set of all pixels belonging to a single blob at time $n$ (part of a blob track). 
\end{itemize}
We compute the centroid $\bm{C}_n$ at time $n$ as follows:
\begin{equation*}
  \bm{C}_n = \frac{1}{|F_n|} \sum\limits_{\bm{x} \in F_n} \bm{x}.
\end{equation*} 
Since columns of a thermal frame are orthogonal to the door opening (Fig.~\ref{fig:sample_thermal}), we use the vertical component $C_n^v$ of centroid $\bm{C}_n=[C_n^h,C_n^v]$ to determine whether a person enters or leaves the room. In particular, we examine whether or not the centroid crosses the mid-line of the frame between two consecutive time instants $n-1$ and $n$. If $C_n^v$ belongs to the upper part of the frame (top $32\times 12$ pixels of the Melexis $32 \times 24$ pixel sensor) whereas $C_{n-1}^v$ belongs to the lower part of the frame (bottom $32\times 12$ pixels) we predict that the person is entering the room. Conversely, if $C_n^v$ belongs to the lower part of the frame whereas $C_{n-1}^v$ belongs to its upper part, we predict that the person is leaving the room. Based on this decision, the people count is updated.

During a hesitant entry/exit or in case of lingering, an event might involve multiple mid-line crossings. We examine the first and last crossings within an event. If the directions of these two crossings are the same, we decide as described above. If the directions differ, we consider this to be a case of lingering and do not update the people count.

\begin{table*}[!htb]
\caption{Details of TIDOS (Thermal Images for Door-based Occupancy Sensing) dataset. Each 32$\times$24-pixel frame was acquired by Melexis MLX90640 sensor at 16 fps. Data was collected by 2 sensors, one over each door of a small classroom.}
   \label{tab:Dataset}
\centering
\begin{tabularx}{\textwidth} { 
  | >{\hsize=0.32\hsize\centering\arraybackslash}X 
  | >{\hsize=0.18\hsize\centering\arraybackslash}X |
   >{\hsize=0.18\hsize\centering\arraybackslash}X |
   >{\hsize=0.18\hsize\centering\arraybackslash}X |
   >{\hsize=1.1\hsize\arraybackslash}X |}
  
 \hline
 Thermal Recording & Number of frames & Number of entries and exits & Initial people count & Challenges ({\it scenario})\\
\hline
 Lecture  & 7,520 & 2 & 9 & Lingering in doorway ({\it only single-person events})\\
\hline
 Lunch Meeting 1  & 37,536 & 25 & 0 & Wearing a coat;
 carrying various items;\newline
 multiple people passing through at the same time\\
\hline
 Lunch Meeting 2 &  9,344  & 8 & 12 & Carrying a backpack ({\it only single-person events})\\
\hline
 Lunch Meeting 3 &  28,128  & 69 & 7 &  Lingering in doorway; wearing a hoodie or carrying a backpack; two people standing in a door and handshaking;\newline
 multiple people passing through at the same time\\
\hline
 Edge Cases &  13,120 & 24 & 6 & Long lingering in doorway; one or two people standing in a door while another person is passing through; \newline
 multiple people passing through at the same time\\
\hline
 High Activity &  22,560 & 133 & 4 & Wearing a hoodie or thick coat; carrying a backpack;  pushing  a chair through doorway; leaning against a closed door; one person standing in a door while another one is passing through; multiple people passing through at the same time\\
\hline
\end{tabularx}
\end{table*}

\section{Experimental Results}

\subsection{Dataset}

We collected a dataset of thermal image sequences using two Melexis MLX90640 32$\times$24-pixel sensors running at 16 Hz mounted above two doors (Fig.~\ref{fig:configuration_sketch}) of a small classroom. Compared to previous research \cite{Thermosense}, \cite{Tyndall_paper} our sensor has a slightly higher spatial resolution, but still a person cannot be visually recognized from the captured data (Figs.~\ref{fig:sample_thermal}, \ref{fig:two people orig}).

Our dataset, called TIDOS (Thermal Images for Door-based Occupancy Sensing), is publicly available\footnote{\href{http://vip.bu.edu/tidos}{\tt vip.bu.edu/tidos}} and includes several types of door activity: single person entering/leaving the classroom, multiple people entering/leaving through the same door, people lingering in the door, people with backpacks, in thick clothing, carrying various items, etc. Details of the dataset are provided in Table~\ref{tab:Dataset}. We manually annotated each frame in the dataset with a number which equals the change in the people count (if any).
Such a change can only occur at the end of an event. During annotation, an event is considered to have ended when a person completely leaves the frame.
We computed the ground-truth people count in the room using our annotations and the initial people count in the room (Table~\ref{tab:Dataset}).

\subsection{Performance Analysis}


We evaluated the performance of our algorithms on TIDOS using the following algorithm parameters: $\alpha = 0.05$ $\sigma = 0.4$ and $\eta = 0.015$ in the RGA model, $\theta P_{F}(T_{n}[\bm{x}])  = 0.015$ (a constant for all $\bm{x}$) and $\gamma = 0.2$ in the MRF-based hypothesis test, and blob-size threshold of $K=L=100$ for both baseline and multi-person algorithms. The values of $\alpha$, $\sigma, \eta, \gamma, \theta P_{F}$ were selected heuristically. However, the values of $K$ and $L$ are motivated by the typical size of a human body's projected image onto the sensor. Based on physical constraints of our setup ($55^\circ \times 35^\circ$ sensor FOV, 2.4m installation height, 1.7m average human height), we concluded that a body's projection typically occupies 200--250 pixels and this agrees with our observation of recorded data. We used 100 as our threshold to avoid misses in case of shorter people, especially children.
%

Since both algorithms estimate transitions in the state of a room (people-count changes), in order to estimate the state of the room (people count) an initial state of the room is needed. In our experiments, we used the true initial people count in each room reported in Table~\ref{tab:Dataset}.
%

\begin{table*}[!htb]
\caption{Performance comparison of the proposed algorithms on TIDOS dataset using three metrics. The lowest values for $MAE$ and $MAE_{PP}$ and the highest value for $CCR_{WCC}$ for each recording are shown in boldface.}
\label{tab:Result_table}
\centering
\begin{tabular}{|c|c|c|c|c|c|c|}
\hline
&
\multicolumn{3}{c|}{Baseline algorithm} &
\multicolumn{3}{c|}{Multi-person algorithm} \\
\hline
      & $MAE$ & $MAE_{PP}$ & $CCR_{WCC}$ & $MAE$ & $MAE_{PP}$& $CCR_{WCC}$ \\
\hline
Lecture & 0.392 & 0.043 &0.500 & {\bf 0.003} & {\bf 0.001} & {\bf 1}\\
\hline
Lunch Meeting 1 & 0.812 &  0.167 & 0.880 & {\bf 0.319} & {\bf 0.065} & {\bf 0.888}\\
\hline
Lunch Meeting 2 & {\bf 0.009} & {\bf 0.001}  & {\bf 0.777} & 0.016 & {\bf 0.001} & {\bf 0.777}\\
\hline
Lunch Meeting 3 & 0.973 & 0.137 &  0.826 & {\bf 0.052} & {\bf 0.007} & {\bf 0.905}\\
\hline
Edge Cases & 0.868 & 0.166  & 0.666 & {\bf 0.548} & {\bf 0.105} & {\bf 0.807} \\
\hline
High Activity & 1.431 & 0.239 & 0.651 & {\bf 0.945} & {\bf 0.158} & {\bf 0.753} \\
\hline
\end{tabular}
\end{table*}

We use three metrics to evaluate the performance of our algorithms. The first two metrics assess the raw people-count estimation performance and are based on Mean Absolute Error ($MAE$).
Our third metric addresses the drift problem, that leads to error accumulation, and temporal misalignments between ground-truth and estimated people-count changes.


\subsubsection{Basic Metrics for Count Estimation}

\begin{figure*}
\centering
\begin{subfigure}[b]{\textwidth}
   \includegraphics[width=0.97\linewidth]{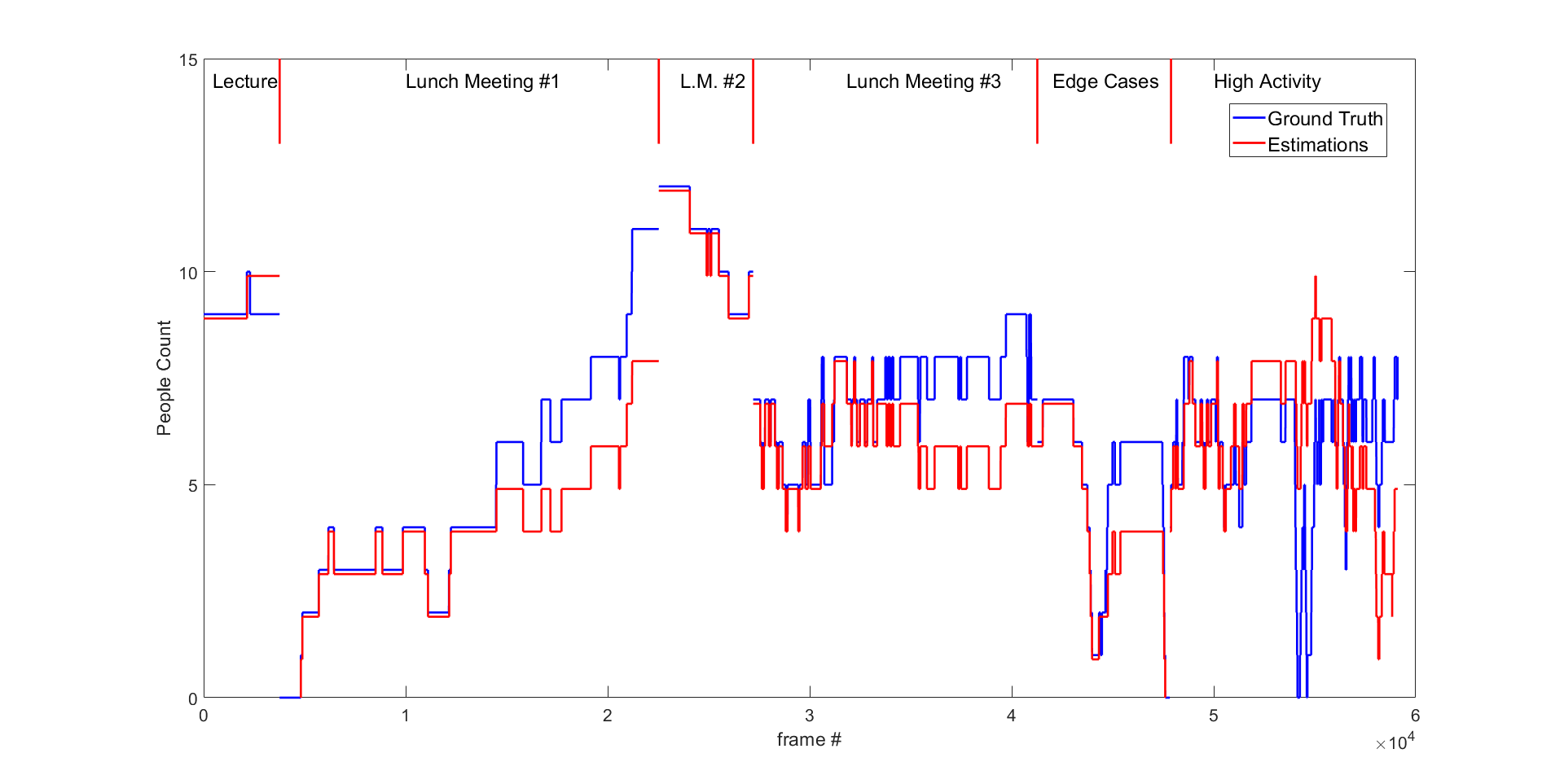}
   \vglue -0.5cm
   \caption{People counts estimated by the baseline algorithm.}
   \label{fig:single_person_DD} 
\end{subfigure}

\begin{subfigure}[b]{\textwidth}
   \includegraphics[width=0.97\linewidth]{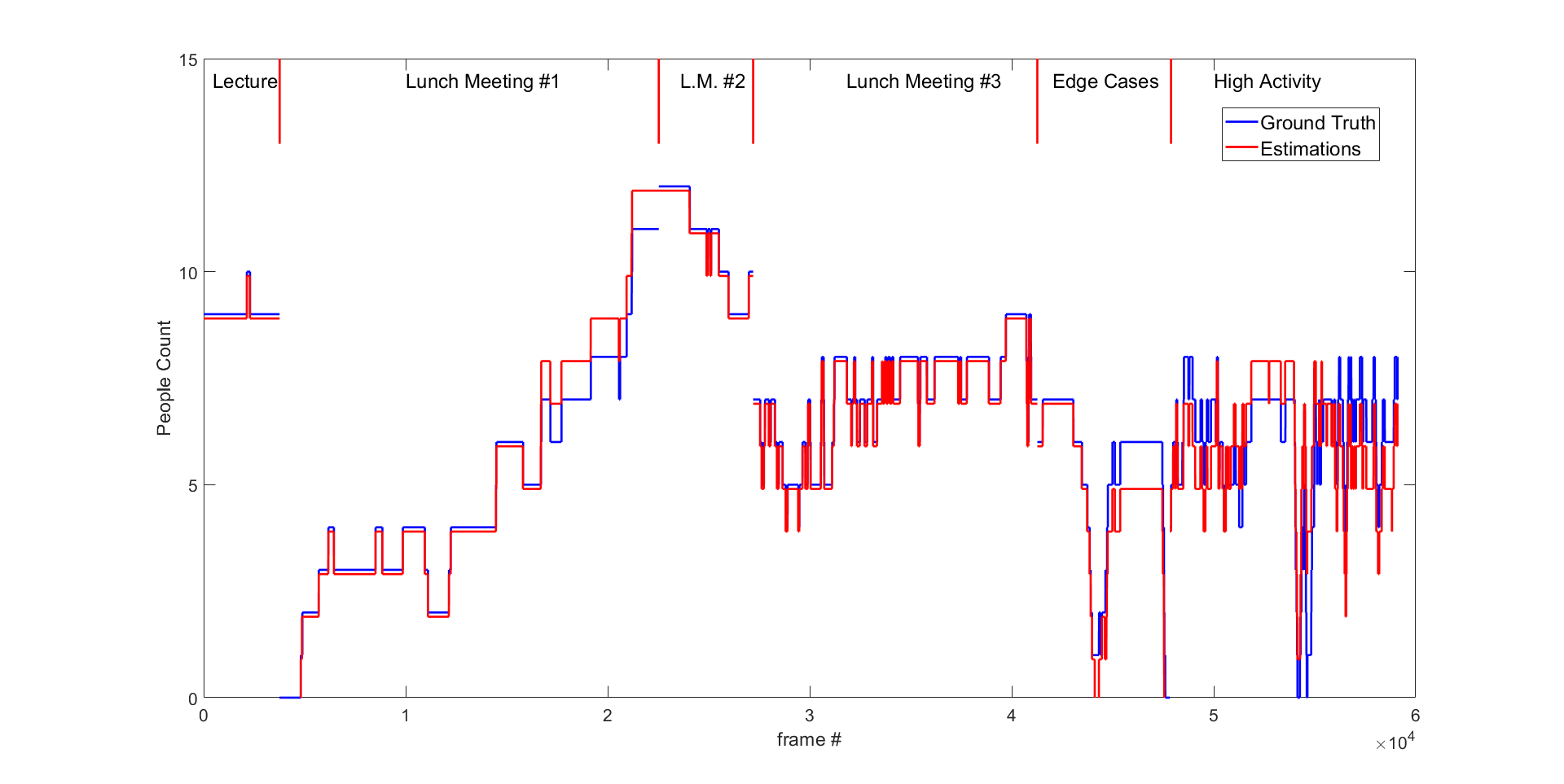}
   \vglue -0.5cm
   \caption{People counts estimated by the multi-person algorithm.}
   \label{fig:multiple_people_DD}
\end{subfigure}
\caption[Two numerical solutions]{True (blue) and estimated (red) people-count plots for the proposed algorithms across all recordings in the TIDOS dataset. To distinguish between the red and blue curves in frames where their values exactly coincide, we added a positive vertical offset of $0.1$ person to the blue curves. Note that since at each time instant two frames are collected (one by each door sensor), the number of frames in these plots is one-half of the total number of frames in Table~\ref{tab:Dataset}.
}
\label{fig:plots}
\vglue -0.4cm
\end{figure*}

Our basic performance metric is the $MAE$ between the true and estimated people counts averaged across all $N$ frames of a thermal sequence. The value of $MAE$ is unaffected by the initial count. However, it scales with the number of people entering/leaving a room which confounds the comparison of $MAE$ values across different occupancy-density scenarios. Thus, we propose another evaluation metric which accounts for the number of people in a room, namely the Per-Person Mean Absolute Error $MAE_{PP}$, 
defined as follows:
\begin{equation}
  {MAE}_{PP} = \frac{\sum\limits_{n=1}^{N} |\hat{y}_n-y_n| } {\sum\limits_{n=1}^{N} y_n },
  \label{eq:MAEPP}
\end{equation}
where $y_n$ and $\hat{y}_n$ are the ground truth and estimate of the number of people in a room at time $n$, respectively, and $N$ is the total number of frames in the recording. While, in principle, the denominator in (\ref{eq:MAEPP}) could be zero, recordings with no people entering/leaving a room are not interesting for algorithm assessment and are absent from our dataset.
We show the performance of our algorithms in terms of $MAE$ and $MAE_{PP}$ in Table~\ref{tab:Result_table} and in terms of frame-wise people count in Fig.~\ref{fig:plots}. Unlike $MAE$, the value of $MAE_{PP}$ is influenced by the initial state of the room since that affects the denominator of Eq.~(\ref{eq:MAEPP}). Moreover, for all recordings in TIDOS, the denominator of Eq.~(\ref{eq:MAEPP}) is larger than $N$, the number of frames in a recording. This causes the $MAE_{PP}$ value to be consistently smaller than the $MAE$ value for the same algorithm applied to the same video.

\smallskip\noindent{\bf Baseline algorithm:} The baseline algorithm has high $MAE$ and $MAE_{PP}$ values for ``Lunch Meeting 1'', ``Lunch Meeting 3'', ``Edge Cases'' and ``High Activity'' recordings. This is due to multiple-person events that the algorithm cannot handle. As expected, the algorithm works well for single-person events as confirmed by low error values for ``Lecture'' and ``Lunch Meeting 2'' recordings.

\smallskip\noindent {\bf Multi-person algorithm:} The multi-person algorithm performs very well on ``Lecture" and ``Lunch Meeting 2" confirming its ability to handle single-person events. It also performs well on ``Lunch Meeting 1", ``Lunch Meeting 3" and ``Edge Cases" recordings that contain multiple-person events. Admittedly, it mishandled one of the multi-person events in  ``Lunch Meeting 1" (Fig.~\ref{fig:multiple_people_DD}, around frame 18,000). The multi-person algorithm does not perform as well on ``High Activity", as it is the most challenging recording in the dataset (see Table~\ref{tab:Dataset}). Not only does ``High Activity" contain the largest number of events, its range of challenges is also widest. Overall, however, the multiperson algorithm significantly outperforms the baseline algorithm in both $MAE$ and $MAE_{PP}$ on all thermal recordings except for ``Lunch Meeting 2'' for which the error is extremely small anyway.

This performance improvement can be also seen in frame-wise people-count plots (Fig.~\ref{fig:plots}). While the baseline algorithm suffers from count drift due to mishandling multiple-person entries/exits (latter parts of ``Lunch Meeting 1'' and ``Lunch Meeting 3''), the multi-person algorithm handles these cases correctly. Clearly, both algorithms have some difficulty with the challenging ``High Activity'' recording but the multi-person algorithm tracks the ground truth more accurately than the baseline algorithm, which is relfected in $MAE$ and $MAE_{PP}$ values.

\subsubsection{Metric Robust to Temporal Misalignments and Error Accumulation}

Despite a very accurate estimate of counts by both algorithms in ``Lunch Meeting 2'' (Fig.~\ref{fig:plots}), their $MAE$ and $MAE_{PP}$ values are not zero. This is due to the fact that although all events have been correctly classified, the timings of a ground-truth event (marked at its completion) and of its estimate may slightly differ. For instance, in the event definition of the multi-person algorithm, a person is considered as ``leaving'' a frame if the associated blob has less than $L$ pixels. However, during our manual annotation a person was considered as out of the frame if s/he left the frame completely. These slight temporal misalignments contribute non-zero values to $MAE$ and $MAE_{PP}$ for a few frames. We can ignore the effects of small temporal misalignments during performance assessment by examining whether the estimated count change occurs within a small temporal window $w$ around the time that the true count change takes place.

Furthermore, $MAE$ and $MAE_{PP}$ apply to people counts and are sensitive to error accumulation because a single miscount could potentially contribute an $MAE$ of 1.0 irrespective of the recording duration $N$. Clearly, a new evaluation metric, resistant to cumulative errors, is needed. Such a metric should focus on \textit{changes} in people counts rather than the counts themselves.

Motivated by these dual considerations, we introduce a new metric, Windowed Count-Change (WCC) Correct Classification Rate ($CCR_{WCC}$), that accounts for both temporal misalignments and error accumulation, and is defined as follows:
\begin{align}
  e_n  &= \min_{-w \leq \delta \leq w} \left|(y_{n+1} - y_n) - (\hat{y}_{n+1+\delta} - \hat{y}_{n+\delta})\right| \nonumber \\
  \delta_n &= \argmin_{-w \leq \delta \leq w} \left|(y_{n+1} - y_n) - (\hat{y}_{n+1+\delta} - \hat{y}_{n+\delta})\right| \nonumber \\
\widehat{\mathcal{N}} &= \bigcup_{m=1}^{N-1}\{m+\delta_m\}, 
\nonumber \\
    CCR_{WCC} &= \frac{ | \{n: (y_{n+1} \neq  y_n) \bigwedge (e_n=0)\}| }
{|\{n: (y_{n+1} \neq y_n) \bigvee (e_n \neq 0) \}| + M} \\
M &= |\{n \notin \widehat{\mathcal{N}}: \hat{y}_{n+1} \neq \hat{y}_{n} \}| \nonumber
\end{align}
%
%
%
%
This metric measures the fraction of frames having count changes in which the estimated count-change equals the true count-change within $\pm w$ frames. However, it ignores the frames for which both the estimated and true changes are zero (no door event) which occur very frequently and would skew the traditional definition of CCR.
$CCR_{WCC}$ is not only resistant to cumulative errors, but also to jitter: even if a prediction is delayed by $\pm w$ frames compared to ground truth, it can still be considered as correct. This metric is essential for applications where misses and false positives need to be minimized, for example monitoring of entryways to a high-security area. A more detailed explanation of $CCR_{WCC}$ can be found on our website.\footnote{\href{http://vip.bu.edu/projects/vsns/cossy/thermal/}{\tt vip.bu.edu/projects/vsns/cossy/thermal}}

However, $w$ needs to be judiciously selected; a large $w$ would unjustly boost $CCR_{WCC}$. We have considered two constraints on $w$, a physically-motivated one and a statistically-motivated one. Given our door setup (sensor's $55^\circ \times 35^\circ$ FOV and 2.4m installation height) and a typical speed of 1.2 m/sec for a person entering/exiting a room, we concluded that this person will be at least partially captured in thermal frames for about 1.3 sec. Therefore, $w$ should be less than 1.3 sec in order to ensure that the person immediately following would not be considered as a potential match within $\pm w$. We have also computed a histogram of time differences between estimated and ground-truth entry/exit times for all events in TIDOS. Over 90\% of these time differences were within 1 sec. Consequently, in all experiments we used $w=16$ frames (1 sec).

The results of Table~\ref{tab:Result_table} show that both algorithms fare equally well in terms of  $CCR_{WCC}$ on ``Lunch Meeting 1'' and ``Lunch Meeting 2'', but the multi-person algorithm clearly outperforms the baseline algorithm by a significant margin on all other recordings. It is also interesting to note that small $MAE$ and $MAE_{PP}$ values need not imply a higher $CCR_{WCC}$ value. Both baseline and multi-person algorithms have lower $MAE$ and $MAE_{PP}$ values for ``Lunch Meeting 2'' than for ``Lunch Meeting 1'', yet their $CCR_{WCC}$ values for ``Lunch Meeting 1'' are much higher than for ``Lunch Meeting 2''. This phenomenon may be partially attributed to the fact that in evaluation metrics such as $MAE$ and $MAE_{PP}$, two errors that occur in opposite directions could cancel out each other. For example, if an algorithm misclassifies one entry event and later misclassifies one exit event, the people count errors due to these two misclassifications will ``cancel'' each other out resulting in zero count errors beyond the second event.

%
%
%
It is clear from Table~\ref{tab:Result_table}, that on ``High Activity"  the multi-person algorithm outperforms the baseline algorithm by a margin of 0.102 in terms of $CCR_{WCC}$ value. 
%
This is a significant improvement because the ``High Activity'' recording has the highest number of entry and exit events and, therefore, a 0.102 fraction of events corresponds to around 13 entries/exits.
Moreover, $CCR_{WCC}$ of 0.753 suggests that three out of four entries and exists were correctly detected and classified within 1 sec of their true occurrence. This is a very solid classification rate for a recording that is mostly composed of very challenging entry/exit scenarios (see Table~\ref{tab:Dataset}). 


\section{Conclusions}

In this work, we developed and systematically studied an overhead virtual tripwire configuration for people counting using a low-resolution thermal sensor. We believe this is the first comprehensive study of its kind encompassing sensor system design and deployment, dataset collection and annotation, algorithm development, design of new performance metrics, and performance evaluation of developed algorithms. The achieved results indicate that typically 80-90\% entry and exit events are correctly classified for scenarios with a wide range of extreme challenges, while in simpler, less-active scenarios even 100\% correct classification can be reached. We hope that this work and the new publicly-available dataset will stimulate further research.

\section{Acknowledgement}
The authors would like to thank Mr.~Yu Xiao, a former Master's student in our group, for developing a dataset annotation tool and for labeling a large portion of the TIDOS dataset.
This research project was supported by the Advanced Research Projects Agency – Energy (ARPA-E), within the US Department of Energy, under agreement DE-AR0000944.

{\small
\bibliographystyle{ieee_fullname}
\bibliography{egbib}
}

\end{document}